# Vehicle-in-Virtual-Environment (VVE)

**Xincheng Cao, Haochong Chen, Sukru Yaren Gelbal, Bilin Aksun-Guvenc and Levent Guvenc**

**Automated Driving Lab, Ohio State University, 1320 Kinnear Rd, Columbus, OH 43212 USA**

**Abstract**

The current approach to connected and autonomous driving function development and evaluation uses model-in-the-loop simulation, hardware-in-the-loop simulation, and limited proving ground work followed by public road deployment of beta version of software and technology. The rest of the road users are involuntarily forced into taking part in the development and evaluation of these connected and autonomous driving functions in this approach. This is an unsafe, costly and inefficient method. Motivated by these shortcomings, this paper introduces the Vehicle-in-Virtual-Environment (VVE) method of safe, efficient and low cost connected and autonomous driving function development, evaluation and demonstration. The VVE method is compared to the existing state-of-the-art. Its basic implementation for a path following task is used to explain the method where the actual autonomous vehicle operates in a large empty area with its sensor feeds being replaced by realistic sensor feeds corresponding to its location and pose in the virtual environment. It is possible to easily change the development virtual environment and inject rare and difficult events which can be tested very safely. Vehicle-to-Pedestrian (V2P) communication based pedestrian safety is chosen as the application use case for VVE and corresponding experimental results are presented and discussed. It is noted that actual pedestrians and other vulnerable road users can be used very safely in this approach.

## 1. Introduction

There have been many well established research developments over the years on active safety and ADAS systems like those in references [1–5]. These have been followed by research on robust and energy preserving control like those in references [6,7] followed more recently by research on autonomous driving like those in references [8–14]. Research work on traffic and energy usage improvement has been reported in references like [15–19] while research work on safety improvements has been reported in references like [9,20–24]. As a result of this and other similar past research, autonomous vehicles that are self-driving and do not need an operator are expected to be available soon. Indeed, there are several limited scale deployments of driverless robo-taxis that are being operated in well structured, geo-fenced areas with warm weather conditions year-round [25]. Unfortunately, most of the development and evaluation of driverless vehicles is taking place on public roads where the other road users are involuntarily taking part in the development of beta level AV software. This approach is in contrast with the well known V-diagram approach of extensive model and hardware in the loop evaluation followed by extensive testing in controlled environments like proving grounds [26]. To elaborate, the usual automotive OEM and supplier development procedure for advanced driver assistance systems involves extensive model-in-the-loop (MIL) simulation followed by hardware-in-the-loop (HIL) simulation and controlled testing in a proving ground to fully develop the system and its software before public road testing using a manufacturer's license plate with several highly attentive and experienced test engineers being present in the vehicle at all times. This final public road testing is carried out to carry out the final tuning of the algorithms and controllers for improved performance in their series production implementation. The well-known V diagram approach of development, evaluation, update and re-evaluation is used during each stage of this well-established development approach [27].

The problem in using this approach in the development and evaluation of autonomous driving functions is that autonomous driving, especially for use inside the city urban environments, relies on scan matching based localization using three-dimensional point cloud maps. Even though highly accurate localization based on RTK GPS is possible, this is not preferred as safe operation requires the autonomous vehicle to localize itself correctly with respect to the road and the surroundings, hence the use of map matching of lidar scans [28]. Well known algorithms like Normalized Distribution Transform (NDT) and Iterative Closest Point (ICP) that are readily available can be used in real time for this map matching based localization [14,29–31]. Unfortunately, map matching based localization cannot be re-created in a proving ground as the surrounding buildings, trees, infrastructure etc that are used as landmarks cannot be replicated. As a partial solution, researchers have built replicas of building blocks within small controlled testing areas [32]. However, this approach only applies to that small building block and the very large variety of surroundings that an AV will encounter in real practice cannot be used in the development and evaluation cycle. Physically changing the building block for each different environment is not feasible as it is very costly and time-consuming. It is also very difficult to re-create the extensive other traffic and weather combinations in this approach. The solution that is currently being used is, thus, doing the final stage of development on public roads which along with being an unsafe approach is also putting the lives of all other road users at risk. This public road development approach is also a very inefficient method since it takes a very long time and very long miles need to be driven to encounter the required rare but extreme situations. The solution proposed in this paper is to replace this unsafe, costly and inefficient public road testing phase with the Vehicle-in-Virtual-

Environment (VVE) method of connected and autonomous driving function development, evaluation and demonstration [33].

The current approach of public road development of autonomous driving functions is illustrated in Figure 1. Note that this is also how an AV operates in the real world. AV sensors including a modem or similar communication device used for connectivity collect data about the surrounding environment. While point cloud lidar data is illustrated in the top left of Figure 1, lidar, camera, radar, GPS and on-board-unit (OBU) modem are generally also used as shown in the bottom left of Figure 1. Sensor data processing and situational awareness algorithms along with decision making are used to generate the higher level trajectory planning or local modifications in order to accommodate other traffic or infrastructure based constraints at a higher level of control while an electronic control unit with CAN connectivity to the throttle, brake and steering actuators implements and executes the lower level controls to follow the required trajectory. The resulting motion of the AV changes its pose (position and orientation) in the driving environment as illustrated in the top right part of Figure 1 where the AV is about to enter a roundabout.

The VVE approach is illustrated in Figure 2. All of the perception, localization and communication sensor data feeds are disconnected using a connection box added to the vehicle. All of the sensor data feeds are instead connected to simulated data from a highly realistic surrounding environment model which can easily be changed, hence the strength of the VVE method in easily being able to use different environments as opposed to the real building block approach. A separate edge computer with a powerful GPU/CPU combination runs the simulated environment in real time and produces the required AV and CV sensor feeds. These are fed to the low level and high level controls in the Control/Processing part of Figure 2 where the high level trajectory planning and local updates and trajectory following controls of the AV work as before but using the simulated sensor data. The low level controls send the actuator commands and the AV moves as before but this time in a large and empty area like a large parking lot or the vehicle dynamics area in a proving ground. The motion of the AV in the large parking lot is tracked using the actual RTK GPS sensor of the vehicle which determines the new pose in the real time simulated environment also. This procedure is illustrated in the top right corner of Figure 2.

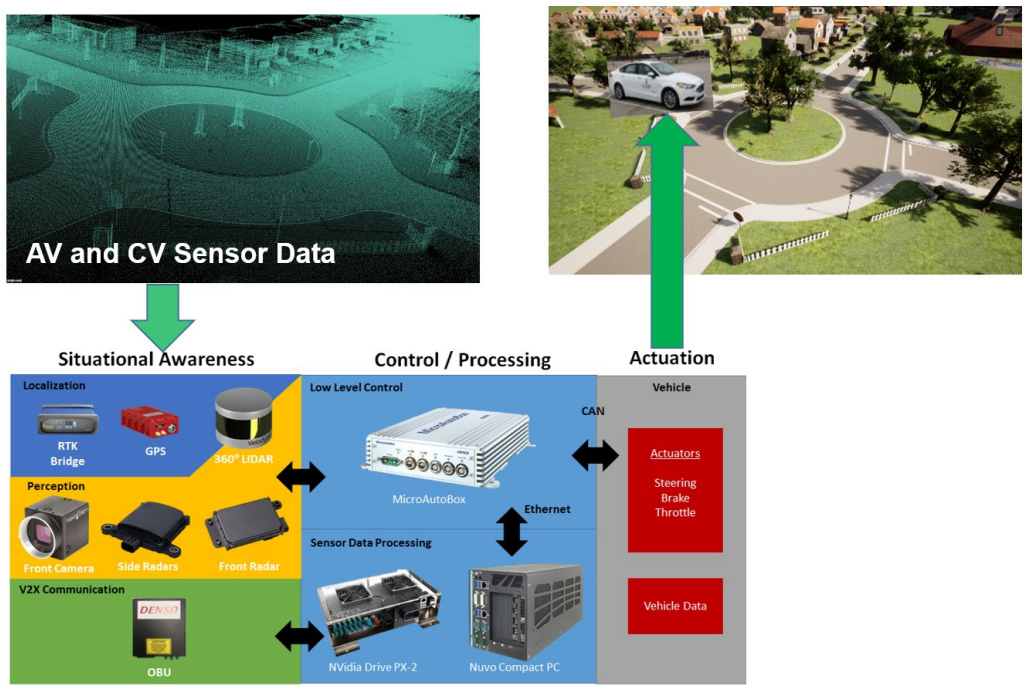

Figure 1. Public road development of AV driving functions.

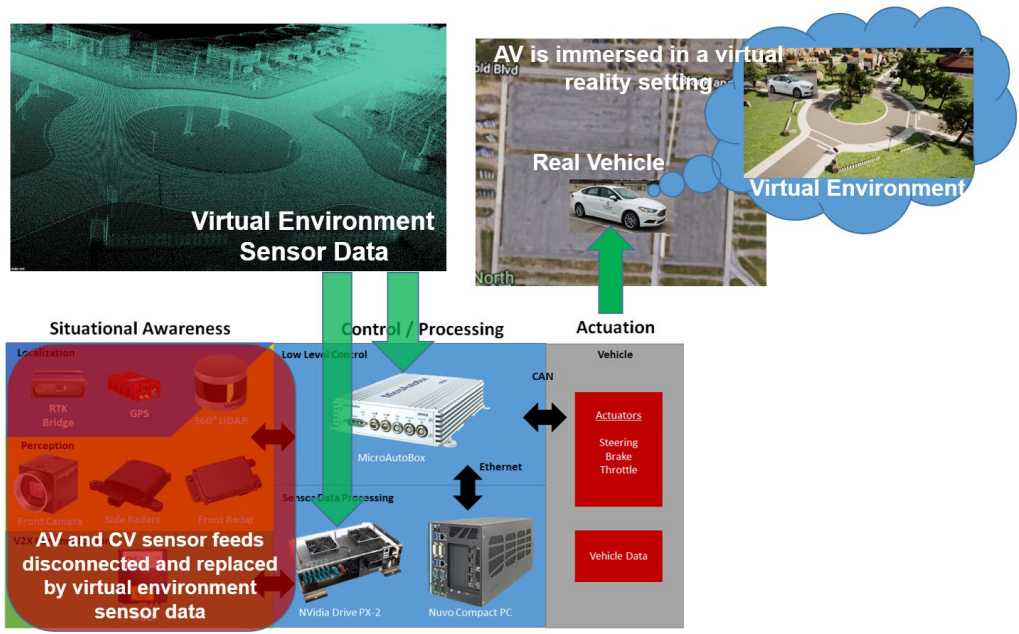

Figure 2. Vehicle-in-Virtual-Environment.

The vehicle itself is, thus, immersed in a virtual reality setting. It is as if the AV now has a VR headset or a holo-lens and is seeing the virtual environment even though it is in a large and flat parking lot or similar test area. Note that it is possible to run the real time environment simulation in the cloud using VVE as a service instead of using edge computing with a very powerful simulation computer in the vehicle. It is also possible to have multiple users share the same virtual environment simultaneously even though they are at different locations, meaning different empty parking lots. This feature allows a very safe method of remotely located teams running and sharing the same AV development and evaluation environment in a safe manner. Other traffic is added realistically using microscopic traffic simulation, also running in real time or as programmed non-player characters in the simulation. While the examples in this paper use Unreal Engine for the simulation environment and CARLA for the AV simulator in Unreal, all of the available 3D environment rendering engines and AV simulators that use them can be used as long as they can run in real time and produce realistic AV and, if needed, CV sensor data. Real CV data for other team members sharing the environment can also be generated by using another modem or communication device. The AV sensor data is converted to the format that is expected by the AV controllers for seamless operation [34]. If the simulation environment is larger than the empty parking lot used, roundabouts are added to the environment to change the direction of the vehicle at the endpoints of the test area and the AV will move to the next building block(s) in the virtual environment.

In order to summarize the motivation and need for the VVE method presented in this introduction, note that a high-fidelity evaluation, development and demonstration method for self-driving that uses the actual autonomous vehicles in a safe but realistic manner is currently not available. This causes high costs and long development times with the risk of unacceptable performance in the form of fatal accidents, too many near misses and too frequent need for operator override. These problems and deficiencies delay the widespread market introduction of self-driving cars and reduce public trust in the global autonomous car industry. Even though this gigantic industry may see a large financial loss due to mistrust in the technology, testing and development is still mainly taking place on public roads. The reasons for this are that the currently used development and evaluation methods rely heavily on pure simulation in the form of model-in-the-loop, hardware-in-the-loop, and vehicle-in-the-loop (still inside a lab environment) forms which keeps the physical localization and perception sensors and a moving vehicle out of the loop. The classical proving ground testing does not have the surrounding building, infrastructure, vegetation, and other traffic environment that are needed to fully test this technology. Attempts at placing replicas of city blocks are useful but can only partially replicate a small, fixed environment and do not solve the problem [32].

The Vehicle-in-Virtual-Environment (VVE) method proposed here takes care of all the problems associated with the current state-of-the-art methods and products by physically driving the actual vehicle in an immersed reality environment while feeding its realistic autonomous driving system sensor signals such that it is fully tested in all possible combinations of environment, other traffic, vulnerable road users, weather conditions and fault situations while being in a very safe actual environment with no collision risk. The advantages of the VVE method over current approaches are illustrated in Figure 3. The VVE approach is expected to be a game changer for the autonomous vehicle industry, legislators, user groups and the public as it will significantly decrease development cost and development time while improving product safety. The cost of the VVE product is also expected to be lower than that of hardware in-the-loop simulators that are widely used for automotive software development and significantly cheaper as compared to proving ground or controlled city block testing. Deployers of technology like Smart Columbus will be able to evaluate a deployment in any geo-fenced urban area they choose while being able to immediately see the effect of unexpected situations in the VVE evaluation. Technology companies will be able to easily demonstrate how their system would operate in a planned deployment site like that in [35] without having to physically go there and spend months of mapping, testing and bug fixing. Certification agencies and local governments will be able to use this tool to fully test vendor technologies before certification and for accident re-construction and analysis.

| Method\Comparison | MIL/HIL | Proving Ground | Building Block | Public Road | VVE |
|---|---|---|---|---|---|
| Implementation | soft/hard | hard | hard | hard | hard/soft |
| Adaptability to Different Scenarios | relatively easy | difficult | very difficult | not possible | easy |
| Vehicle Model | high fidelity | real vehicle | real vehicle | real vehicle | real vehicle |
| Safety | safe | controlled experiment necessary | controlled experiment necessary | not safe | safe |
| Cost | high | very high | very high | very high | moderate |
| Repeatability | high | high | high | low | high |
| Time to Implement | moderate | long | long | very long | moderate |

Figure 3. Advantages of VVE over current methods.

After this introduction in Section 1, the rest of the paper focuses on an application use case to illustrate how the VVE method works. A more detailed explanation of how the VVE is implemented is presented in Section 2 using basic manual driving and path following inside a virtual environment with a real AV. This is followed in Section 3 by the discussion of the application use case on evaluation of pedestrian safety using Vehicle-to-Pedestrian (V2P) communication and the presentation of the test results in Section 4. The paper ends in Section 5 with conclusions and recommendations for future work.

## 2. The VVE Method

Our current VVE architecture implemented in the ego AV is illustrated in Figure 4 which shows how motion in the actual empty parking lot and motion in the corresponding virtual world are correlated with each other. As noted before, while our current architecture uses an Unreal Engine rendering of the virtual test environment and the CARLA AV simulator, any of the currently available three dimensional surrounding environment modeling and AV simulation tools can be used. At the beginning of the VVE run, the vehicle in the empty parking lot is placed at a desired reference position corresponding to the initial position of the virtual vehicle in the virtual environment. Both vehicles start at the same orientation. The real vehicle motion changes in the empty parking lot are then recorded and translated into the corresponding motion changes in the virtual environments such that the virtual vehicle moves by the same amount in the virtual world that the real AV is immersed in. At each new position and orientation in the virtual environment, sensor data is collected in the simulation computer and sent to the AV computer system.

The actual vehicle used in this paper is shown in Figure 5. The simulation computer, the perception sensor computer, the low level control computer and the GPS processing unit are shown in the trunk of this vehicle in Figure 5. The dSpace microautobox unit is a generic electronic control unit with CAN and Ethernet connections and acts as the low level controller. The calculated trajectory or trajectory modification is tracked within the steering and speed controller implemented in this low level control unit which sends the drive-by-wire CAN commands for throttle, brake and steering to the AV drive-by-wire interface. An RTK GPS unit is used to keep track of position and orientation changes which are then conveyed to the virtual world simulation to read perception sensors at the new virtual world location. Figure 6 shows the planned path in the virtual world on the right and the actual AV in the parking lot that is immersed in that environment and following that path on the left.

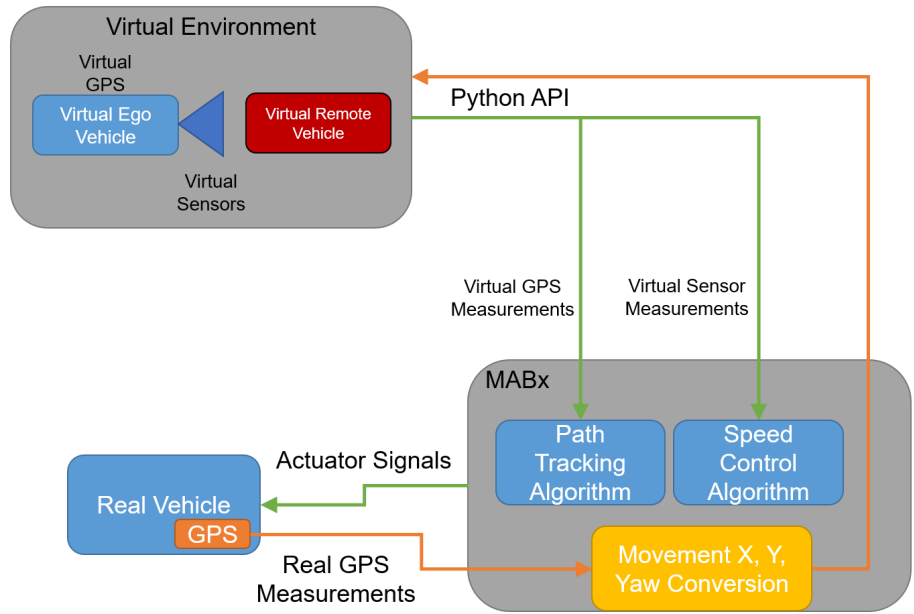

Figure 4. Correlation of motion in the real and virtual environments for a simple trajectory tracking application.

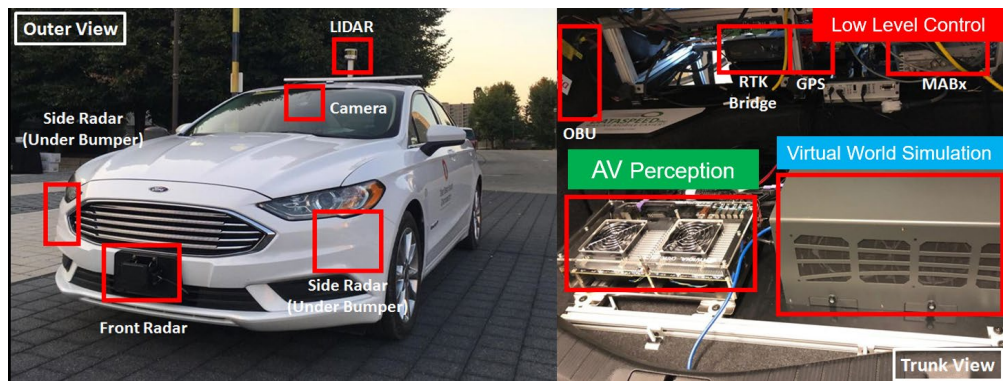

Figure 5. Research level AV used and the relevant components in its trunk.

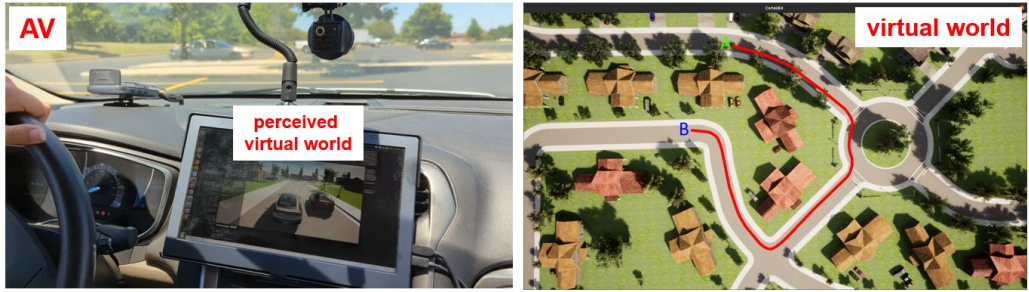

Figure 6. Actual AV in parking lot and planned path in the virtual world.

## 3. Pedestrian Safety Using V2P

The application use case for VVE that is used in this paper is pedestrian safety using V2P communication. The V2P vulnerable road user safety mobile phone app developed in our earlier work in reference [21] is used here for the communication between the AV and pedestrian. Development of pedestrian and vulnerable road user safety systems in public roads is not recommended due to safety issues. The alternative approach of using a mannequin on a movable platform for controlled testing in a proving ground is useful but is very limited in scope considering the many different situations that occur in real life when AVs and CVs encounter and interact with vulnerable road users. The VVE method is an excellent choice here because, along with software based vulnerable road users, it is also possible to use real vulnerable road users that share the same virtual environment and move at displaced and safe locations while the AV in the empty parking lot will perceive them to be on a collision risk path. This section, therefore, starts with Vehicle-to-Everything (V2X) and V2P communication and proceeds with how to implement the V2P based VVE testing.

Vehicle connectivity research has seen rapid advancement in recent years. From the perspective of safety, connectivity can handle some traffic scenarios that are traditionally challenging, such as pedestrian motion detection under NLOS (no-line-of-sight) conditions. References [21,36–38] provide some examples of pedestrian motion tracking and collision risk assessment implementations through cellular, Bluetooth or Wi-Fi connections for example. There are two groups of technologies being used currently for V2X communication: Wireless-Local-Area-Network (WLAN) based solutions and cellular-based solutions (C-V2X). WLAN-based technologies are based on the IEEE 802.11p standard [39]. Reference [40] offers a performance evaluation of IEEE 802.11p and IEEE 1609 WAVE (another standard built upon IEEE 802.11p) standards in sense of capacity and delay and concludes that the traffic prioritization schemes work well and stable connections in high density traffic is possible. The most notable technology used in this branch is Dedicated Short Range Communication (DSRC), where direct communication among vehicles and infrastructure can be established. It operates in the 5.9 GHz band with a bandwidth of 100 MHz in the U.S., and its devices have an operation range of 1 km [41].

Cellular-based technologies are another popular area of V2X connectivity. These technologies are developed under 3GPP (3rd Generation Partnership Project) and include a wide range of protocols such as GSM (Global Mobile Communications System)/2G, UMTS (Universal Mobile Telecommunications System)/3G, LTE (long-term evolution)/4G, 5G NR (5G New Radio)/5G cellular networks with 6G on the way. Reference [42] provides an overview of 3GPP standards and offers technical comparison between 3GPP functionalities and IEEE 802.11p standards. Reference [41] also provides a brief introduction of GSM network topology. In recent years, LTE and 5G based solutions are most explored as they offer significant advantages over previous generation cellular networks. Reference [43] tests and compares LTE and 5G NSA (non-standalone) network under V2X application and observes significant better performances of 5G NSA compared to LTE in sense of response time and packet loss.

Apart from the aforementioned two main groups of solutions, some other connectivity options exist. Wi-Fi is a wireless connection protocol based on earlier variations of IEEE 802.11 standard, but it is not a suitable option for V2X applications due to its varying data rate under different conditions [41]. ZigBee is a communication scheme based on IEEE 802.15.4 [39] and is another possible alternative for V2X connectivity. Reference [41] tests the handshake time of different ZigBee channels. Bluetooth is another short-range wireless communication option and is explored by many recent works. Reference [44] describes an Android application that tracks real-time vehicle motions and uses Bluetooth to transmit information received on DSRC devices to connected mobile phones. References [45] and [41] analyze the handshake time of Bluetooth connection under noisy Wi-Fi conditions. Reference [21] introduces a mobile phone application that broadcasts PSM (personal safety messages) between vehicle and pedestrian via Bluetooth low energy connection using the advertising mode. An extension of this last Bluetooth BLE communication app between two mobile phones will be used here as it has performed very well in recent deployments. The pedestrians or vulnerable road users run the app in their mobile phones which broadcasts their location information using PSM to nearby vehicles where another mobile phone or Bluetooth device running the software listens to this information and uses it to determine collision risk with the pedestrian or vulnerable road user. If the collision risk is high and the vehicle and pedestrian or vulnerable road user are close, the AV applies autonomous braking to avoid an accident. This V2P communication is illustrated in Figure 7. It should be noted that C-V2X and over-the-cloud connectivity can also be used to obtain similar results and can be tested using the VVE method. Figure 8 shows the VVE implementation architecture for developing, evaluating and demonstrating V2P based vulnerable road user safety. Experimental results are presented and discussed in the next section.

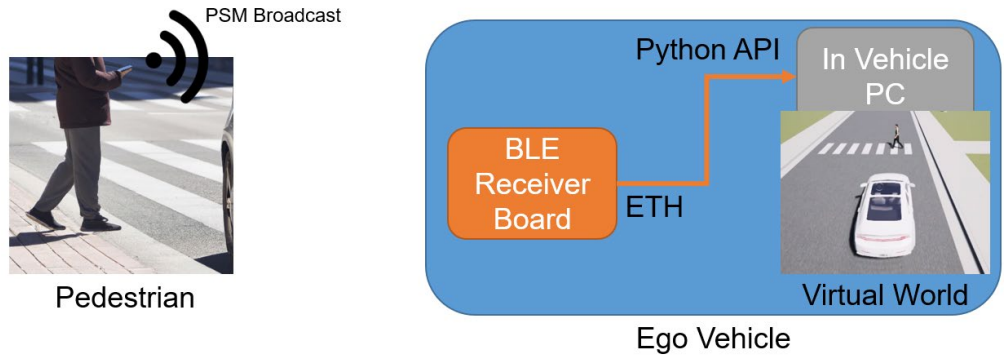

Figure 7. Mobile phone BLE based communication between vulnerable road user and AV.

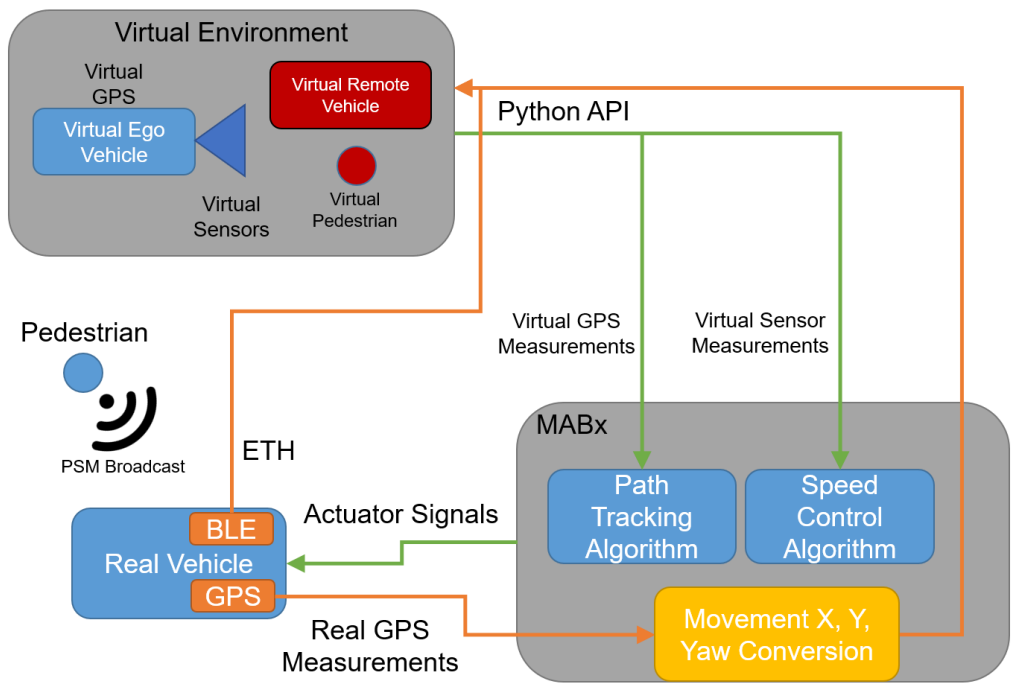

Figure 8. V2P based vulnerable road user safety implementation of VVE architecture.

## 4. Experiments of Pedestrian Safety Using V2P

This section presents a proof-of-concept demonstration of V2P functionality in a virtual environment using pedestrian safety through V2P communication as an example. The Carla simulator and Unreal Engine are selected as the AV simulator and environment modeling tool, respectively, for this demonstration. The collision risk estimation routine that is used is illustrated in Figure 9 [21]. The vehicle and pedestrian headings are first compared, and if their future paths intersect one another, the intersection point becomes the potential collision point. A collision zone is, then, established around this collision point, in this case as a rectangular area of size 6m x 6m. Based on the current heading and speed of the vehicle and the pedestrian, Time-To-Zone (TTZ) can be calculated separately from the perspective of both the vehicle and the pedestrian. The two TTZ values are then compared to each other, and if their difference is small enough, collision is deemed highly probable as the vehicle and the pedestrian are expected to arrive at the collision zone at the same time. For this implementation, the TTZ difference is compared to a threshold value Ts, in this case chosen as 1.5 sec, to determine if the situation is potentially dangerous. Once a situation is deemed dangerous, automatic braking will be applied to the vehicle to avoid possible collision. To accommodate different situations, a three-level severity classification is implemented as shown in Figure 10. Once the TTZ difference is within the chosen threshold, the TTZ value for the vehicle is used to determine the severity level of the possible collision. In general, a smaller TTZ value indicates a shorter headway time to collision, and hence requires harder braking. In this case, the TTZ threshold value used to differentiate level 1 and 2 severity is chosen as 2.3 sec, and the TTZ threshold value used to differentiate level 2 and 3 severity is chosen as 1.5 sec. It should again be noted that the threshold values and collision zone sizing can be easily modified to accommodate various settings such as different vehicle dynamic models and road conditions.

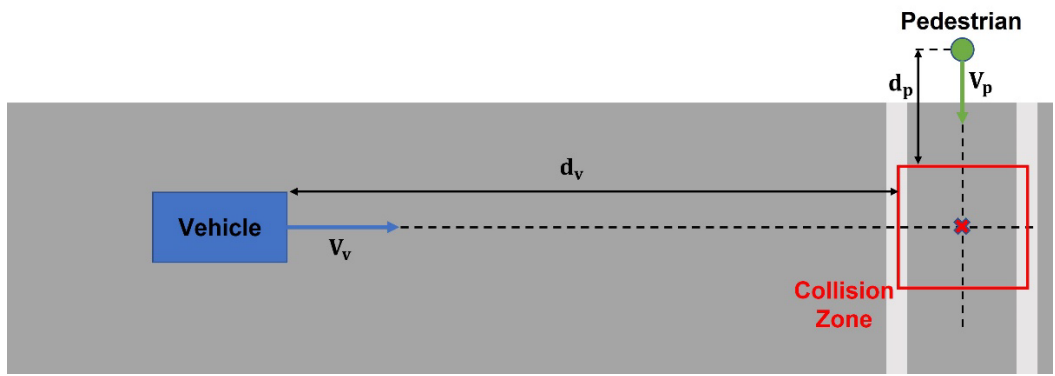

Figure 9. Collision risk estimation.

| Severity Level 1 | Sufficient distance to collision zone, Mild braking necessary. |
| Severity Level 2 | Close to collision zone, Moderate braking necessary. |
| Severity Level 3 | Very close to collision zone, Emergency braking necessary. |

Figure 10. Severity levels.

We present a traffic scenario as displayed in Figure 11. The ego vehicle approaches an intersection, where a pedestrian intends to cross. Another vehicle is parked at the intersection in a neighboring lane, blocking the line-of-sight (LOS) between the ego vehicle and the pedestrian. This is a typical traffic case, and the no-line-of-sight (NLOS) condition makes it difficult for the ego vehicle's onboard sensors to detect the pedestrian. In Figure 12, simulation results are presented for the worst-case scenario, where V2P connection is not implemented, and the pedestrian decides to quickly run across the intersection as the ego vehicle approaches, necessitating emergency braking. It can be observed, however, that the ego vehicle fails to decelerate for the crossing pedestrian who it does not detect, and collision becomes imminent.

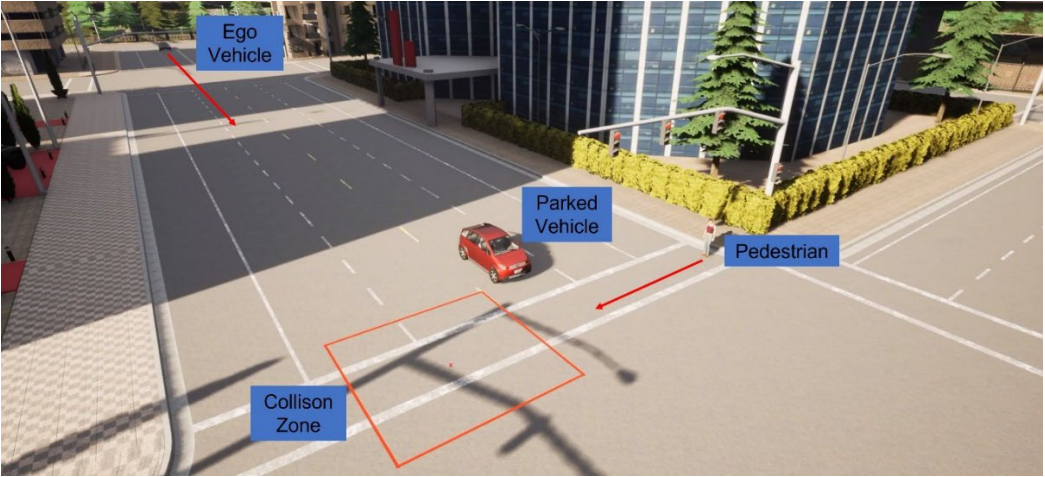

Figure 11. NLOS intersection scenario with suddenly darting pedestrian.

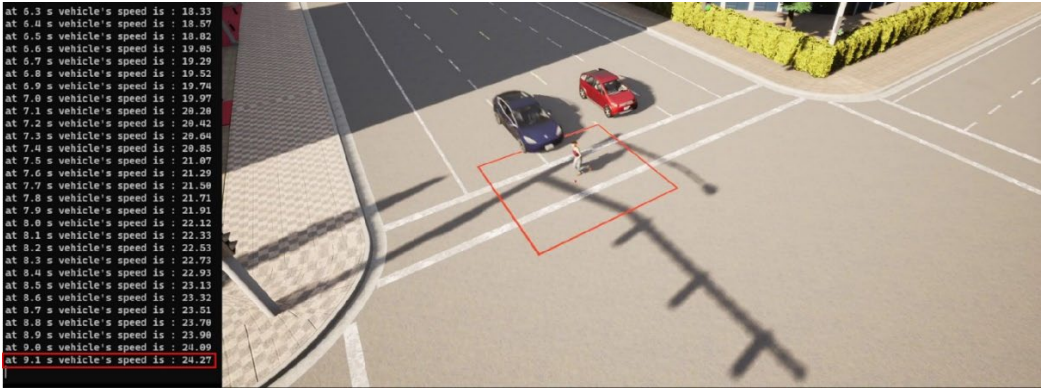

Figure 12. Automatic braking not engaged, collision imminent.

We then implement the V2P communication based autonomous braking scheme introduced above and run the experiment again, the results of which are presented in Figure 13. It can be observed that the ego vehicle begins to engage the brake before it is able to establish a LOS with the pedestrian and is able to eventually come to a stop before colliding with the pedestrian. In this case, a level 3 severity is needed, and maximum braking is applied to avoid collision. In order to demonstrate the functionality of the three-level severity design, two more cases are experimented. In the case displayed in Figure 14, the pedestrian walks slowly across the intersection and the ego vehicle has ample time to react. As a result, only a level 1 severity is needed, and the ego vehicle only needs to apply minor braking to stop and avoid collisions. In the case displayed in Figure 15, the pedestrian runs across the intersection while the ego vehicle is still somewhat far away, allowing the ego vehicle to avoid collision by applying moderate braking action triggered by a level 2 severity classification.

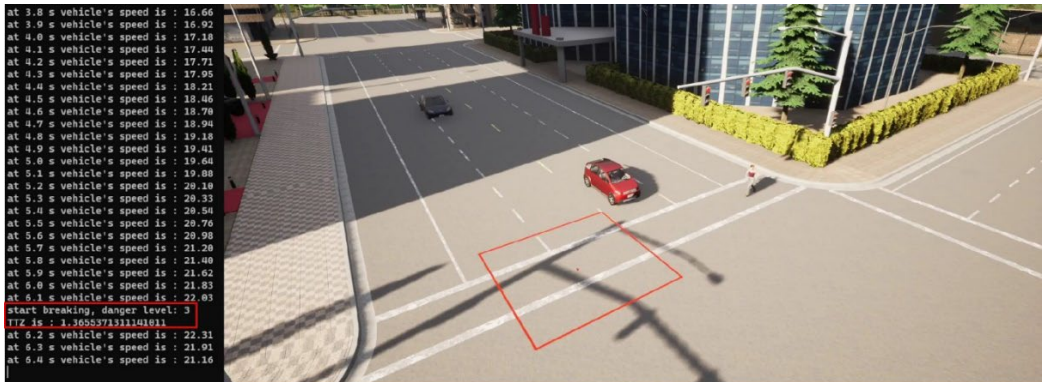

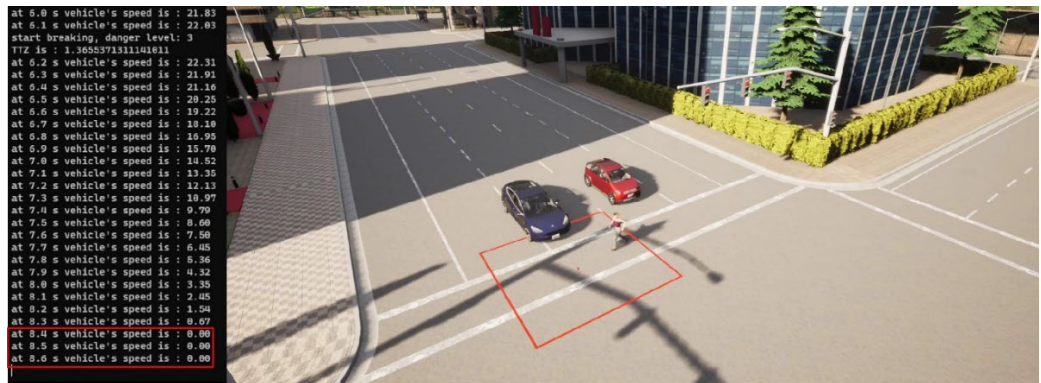

Figure 13. (a) Automatic braking engaged, severity level 3; (b) Vehicle stopped, collision avoided.

In the VVE experiments presented above, the vehicle and pedestrian are at two close but different locations with no possibility of a real collision as illustrated in Figure 16. A mobile phone is placed in the vehicle, and another mobile phone is in the pedestrian's possession and they both use a V2P communication app that sends PSM data of the pedestrian to the vehicle. The vehicle is in an open space, presumably a parking lot, so that it can maneuver, while the pedestrian is at another safe location. Both mobile phones are connected to the same Carla environment, and their sensor data are fed into the environment. Collision risk is calculated in the environment and the appropriate level of braking command is sent to the vehicle that facilitates the braking action in the parking lot. As a result, it is possible to realistically and safely test different vehicle and pedestrian interactions including dangerous ones.

## 5. Conclusions and Recommendations for Future Work

The VVE method was introduced in this paper as a safe, efficient and low cost method of developing, evaluating and demonstrating connected and autonomous driving functions.

The widespread use of VVE is expected to replace the current unsafe and time consuming approach of public road development of AV driving functions. A path following task was used to illustrate how the method works. V2P communication based vulnerable road user safety was chosen as the application use case in this paper and VVE runs were used to demonstrate how the method can safely be employed with real pedestrians and an AV in a parking lot that are all immersed in the same realistic, three dimensional environment. Results for non-line-of-sight pedestrians including a sudden darting pedestrian were used in the evaluations demonstrating the efficacy of the VVE method. It is recommended that future work concentrate on more application use cases to demonstrate the full potential of the VVE method and help with its widespread adoption.

Some other references that could be useful for future work related to this method or useful in relation to the function/control etc to be developed, evaluated or demonstrated can be found in references [46-50].

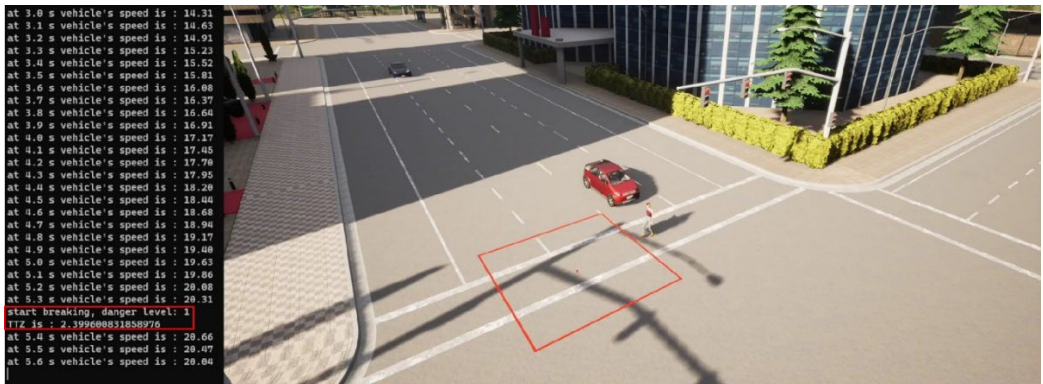

(a)

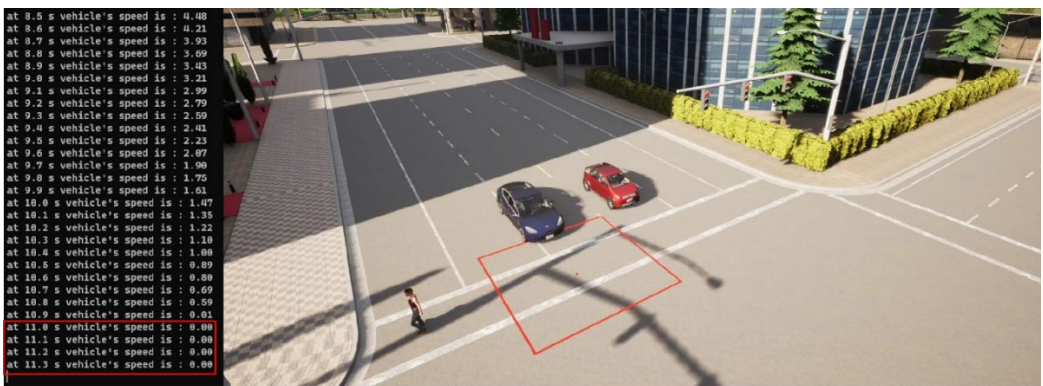

(b)

Figure 14. (a) Automatic braking engaged, severity level 1; (b) Vehicle stopped, collision avoided.

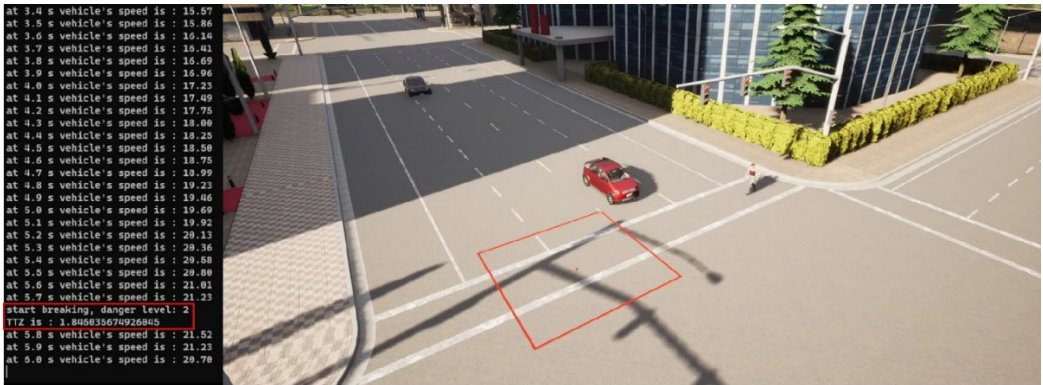

(a)

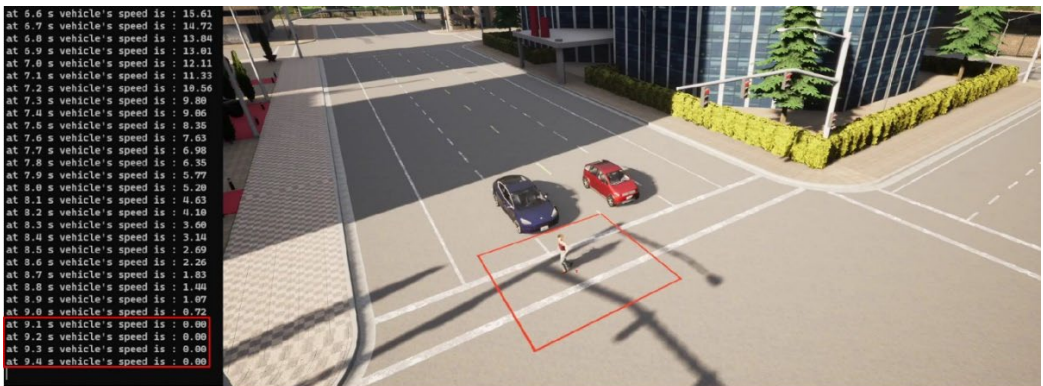

(b)

Figure 15. (a) Automatic braking engaged, severity level 2; (b) Vehicle stopped, collision avoided.

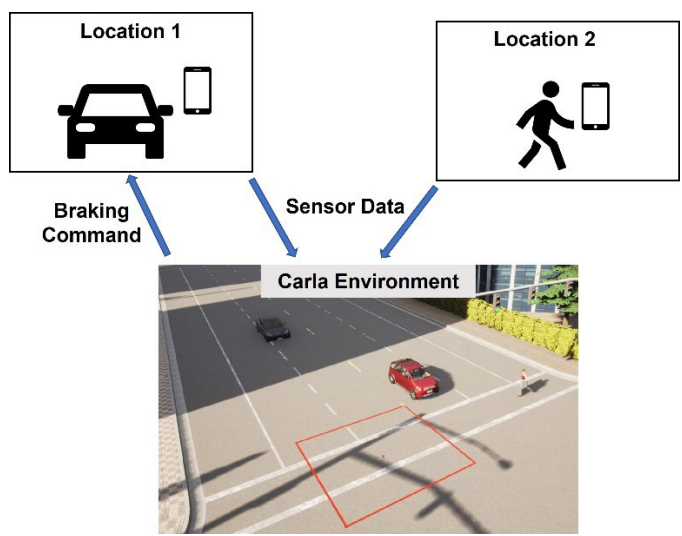

Figure 16. Experimental setup.